\documentclass[review]{elsarticle}

\usepackage{lineno,hyperref}
\modulolinenumbers[5]
\usepackage{amsmath}
\usepackage{multirow}

\journal{Pattern Recognition}

\begin{document}

\begin{frontmatter}

\title{Deep Domain Generalization with Feature-norm Network}





\author{Mohammad Mahfujur Rahman \corref{cor1}} 
\cortext[cor1]{Corresponding author: 
  Tel.: +61416453032;}  
\ead{m27.rahman@qut.edu.au}
\author{Clinton Fookes}
\author{Sridha Sridharan}

\address{Signal Processing, Artificial Intelligence and Vision Technologies (SAIVT), Queensland University of Technology (QUT), Brisbane, QLD 4000, Australia}

\begin{abstract}
In this paper, we tackle the problem of training with multiple source domains with the aim to generalize to new domains at test time without an adaptation step. This is known as domain generalization (DG). Previous works on DG assume identical categories or label space across the source domains. In the case of category shift among the source domains, previous methods on DG are vulnerable to negative transfer due to the large mismatch among label spaces, decreasing the target classification accuracy. To tackle the aforementioned problem, we introduce an end-to-end feature-norm network (FNN) which is robust to negative transfer as it does not need to match the feature distribution among the source domains. We also introduce a collaborative feature-norm network (CFNN) to further improve the generalization capability of FNN. The CFNN matches the predictions of the next most likely categories for each training sample which increases each network's posterior entropy. We apply the proposed FNN and CFNN networks to the problem of DG for image classification tasks and demonstrate significant improvement over the state-of-the-art.
\end{abstract}


\end{frontmatter}


\section{Introduction}
Deep neural networks (DNNs) \cite{NIPS2012_4824} achieve excellent performance for classification tasks in computer vision when the training and test data are collected from an identical distribution. However, this supposition cannot consistently be true in practice due to various factors including viewpoint changes, illumination variation and background noise that can cause domain shift. DNNs are known to be vulnerable to such domain shifts. The approaches for solving the domain shift issue can be broadly classified into two categories: domain adaptation (DA) 

\begin{figure*}
\begin{center}
\includegraphics[width=1.0\linewidth]{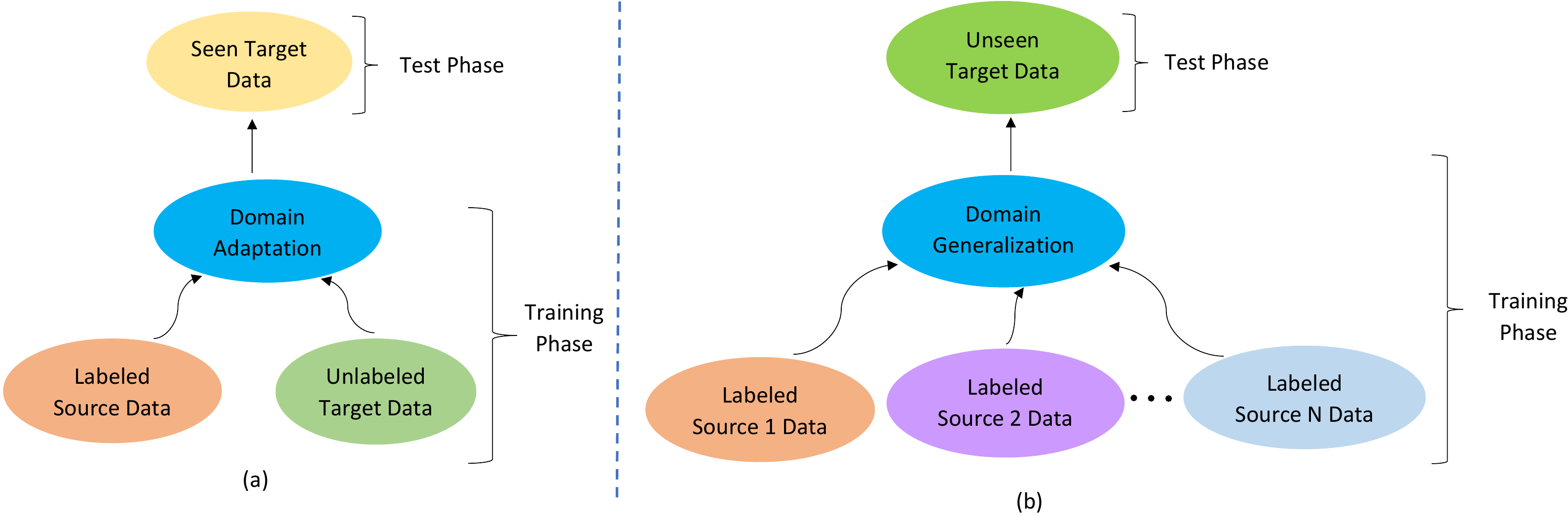}
\end{center}
   \caption{Comparison between domain adaptation and domain generalization. (a) Domain adaptation can access the unlabeled target data during training phase. (b) Domain generalization cannot access the target data during training phase.}
\label{fig:diff_DA_DG}
\end{figure*}

\cite{pmlr-v37-ganin15,open_2017_ICCV,Cao_2018_CVPR,Rahman2020,chen2020deep,liang2019exploring,mahapatra2020training,chen2020adversarial} and domain generalization (DG) \cite{DBLP:journals/pami/GhifaryBKZ17,Muandet:2013:DGV:3042817.3042820,DBLP:conf/iccv/GhifaryKZB15,8053784,7437460,Li_2018_CVPR,DBLP:journals/corr/abs-1807-08479,8237853,RAHMAN2020107124,rahman2019multi,matsuura2020domain,ryu2019generalized,qiao2020learning,zhang2020unsupervised,mahajan2020domain,zhou2020learning,du2020learning,wang2020learning,matsuura2020domain,tseng2020cross,ryu2019generalized,qiao2020learning}. DA methods access the target data during training  to train the model using the combination of labeled source data and unlabeled / sparsely labeled target data. In contrast, DG is a far more challenging task to solve as it implies no access to the target data during training. With training on the labeled data of multiple source domains, the model can be applied to any unseen target domain without adaptation stage in testing.  In real-world situations, DG methods are more effective than DA techniques, because DG techniques do not concentrate on adapting the source data to the specific target domain but seek to generalize to unknown target domains. DG considers the situation where the labelled data is accumulated from multiple source domains so that the trained model will be able to manage new domains without the adaptation stage. 


Most of the existing DG methods achieve the generalization ability by extracting features from the pre-trained DNNs or using handcrafted features. The shallow architecture based DG approaches \cite{DBLP:journals/pami/GhifaryBKZ17,Muandet:2013:DGV:3042817.3042820,DBLP:conf/iccv/GhifaryKZB15,8053784,7437460} cannot fully exploit the benefits of deep learning as a linear transformation is used after feature extraction. We contend that learning the domain agnostic feature transformation in an end-to-end approach straightaway from the actual image will assist to achieve even more desirable performance.

\begin{figure*}
\begin{center}
\includegraphics[width=1.0\linewidth]{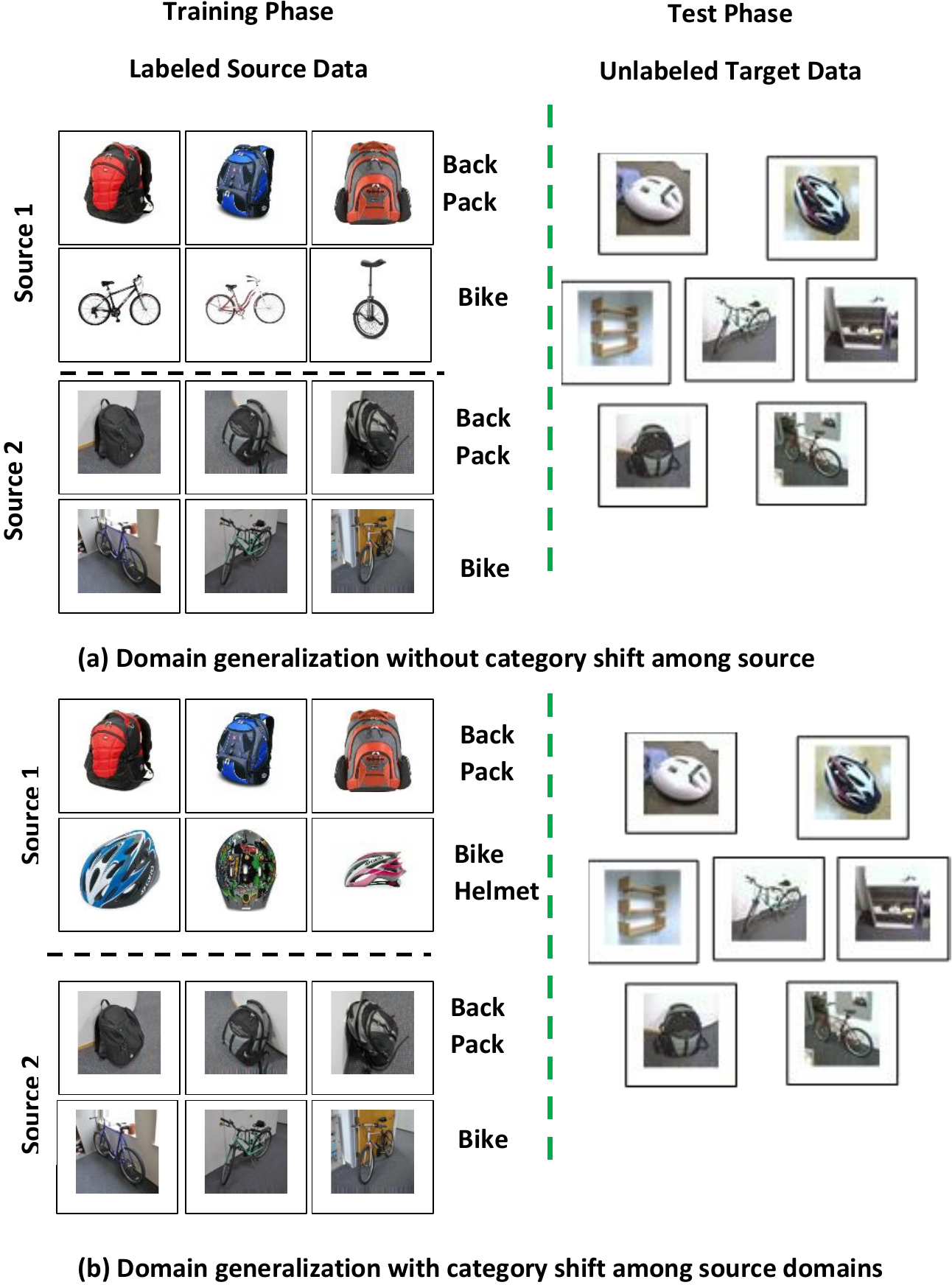}
\end{center}
   \caption{The difference between the conventional setting in domain generalization and the category shift setting in domain generalization. (a) Conventional domain generalization methods assume the source domains share the same label space. (b) Domain generalization methods with category shift assume the source domains have uncommon categories between each other.}
\label{fig:diff_DA_DG}
\end{figure*}

Another major limitation of current approaches to increase the generalization of a model is the assumption of an identical label space among the available source domains. Recently, Li et al. \cite{Li_2018_CVPR} introduced an adversarial autoencoder network for DG  where the discrepancy is minimized using maximum mean discrepancy (MMD). Li et al. \cite{Li_2018_ECCV} proposed a deep DG framework where the divergence of the source domains is reduced by the adversarial learning. They assume the same label space exists among all the source domains. When the source domains have some uncommon classes, if all the distributions are matched among the source domains, existing DG approaches fail to yield  domain invariant features. This is due to negative transfer learning which is an undesirable consequence of transfer learning \cite{DBLP:journals/tkde/PanY10} where the final performance is worse than a classifier trained on the source domains with a supervised classification loss. Therefore in the case of category shift among the source domains, the unshared classes should not be matched. As shown in Figure \ref{fig:diff_DA_DG}, DG with category shift in the source data is more challenging than DG without category shift, since unshared source classes will result in negative transfer when minimizing the discrepancy among the domains. However, in practice, it is usually difficult to obtain multi-source domains with an identical label space. Previous DG approaches were therefore limited to using only convenient datasets to obtain a domain agnostic model. As a consequence, their generalization ability  have been severely limited. For example, ImageNet dataset has $1000$ classes and Caltech-256 dataset has $256$ classes. If we compare all the statistics using traditional adversarial learning or moment matching techniques in the settings of DG, we will face the problem of negative transfer leaning.


To address the above mentioned limitations of traditional DG techniques, we introduce an end-to-end novel feature-norm network (FNN) for DG where we assume the source domains have category shifts. In our approach, the whole statistics from one domain are not compared with other domains to reduce domain discrepancy. Rather, we reduce domain disparity by enlarging the feature-norm of every individual example. We argue that the domain discrepancy among the source domains largely depends on the smaller norm features. We therefore surmise that if we can come up with a strategy to enlarge the feature-norm, the domain discrepancy among the source domains will be reduced. We introduce a novel approach based on this perception, as an alternative to comparing the whole statistics of a domain to others.

Our proposed approach is able to perform in both scenarios where either the source domains share the same label space or share a partial label space. Feature-normalization has recently been shown to render a substantial advancement in performance \cite{DBLP:journals/corr/WangXCY17,DBLP:journals/corr/RanjanCC17} in other applications such as face verification. As outlined in \cite{DBLP:journals/corr/abs-1803-00130,9009009}, a motivation behind such approach is that low norm features play a less informative role in the presence of domain discrepancy. Thus, we impose large norms on the task-specific feature embeddings to accelerate higher transferable possibility. Xu et al. \cite{9009009} suggested that the domain shift between the source and target domains relies on the excessively less informative features with the smaller norm and introduced a DA approach. In contrast, we solve the DG task instead of DA by minimizing the distance among the source domains during training phase. In DG, the target data is totally unseen in the training phase but used in the testing phase. The difference between the approach of \cite{DBLP:journals/corr/abs-1803-00130} and our proposed method is the adaptive radius control. In our method, during each iteration, we enlarge the feature norm according to the feature norms of the last iteration computed by the previous parameters of the model while the radius of the hyper-sphere is fixed in \cite{DBLP:journals/corr/abs-1803-00130}.

Our proposed FNN is further extended by the proposed collaborative feature-norm network (CFNN) for DG. In CFNN, both networks predict the true class for each training sample with a supervised classification loss, but the estimation of the next most likely classes are different as the networks start with a distinct initial state. CFNN merges the collective estimation of the next most probable categories. The matching of each training sample in other most likely categories increases each network's posterior entropy \cite{a03d434cfce142c3b10c35ae1062865d} that boosts them to assemble better generalization to test data. Moreover, CFNN improves the generalization capability since CFNN have lower variance error compared to FNN. The contribution of this work is three-fold:
\begin{itemize}
\item We propose an end-to-end feature-norm network that enables generalization to the unknown target domain. 
\item We further improve our proposed feature-norm network by introducing collaborative learning in the context of DG. CFNN improves the generalization capability to the test data by matching the predictions of the next most likely categories.
\item  Our proposed approach is independent of the association of the label spaces among source domains, hence it can handle the comparatively harder scenarios where the source domains have some unshared categories. 

\end{itemize} 

The reaming sections of the paper are structured as follows. In section 2 we discuss related works, in section 3 we illustrates the proposed approach, in section 4 we reports the results and finally section 5 concludes the paper.  

\section{Related work}

In computer vision, DA has experienced a remarkable research focus in recent years and it has been examined under many different settings: closed-set \cite{pmlr-v37-ganin15}, open-set \cite{open_2017_ICCV}, partial \cite{Cao_2018_ECCV}, single source domain \cite{pmlr-v37-ganin15} or multiple source domain \cite{NIPS2018_8075}. However, in DA settings these approaches require either labeled target data or unlabeled target data during the training procedure. Both DA and DG methods are used to mitigate the domain shift problem. The dissimilarity between DA and DG is the accessibility of the target data in the training phase. The advantage of DG over DA is that in DG if we have a new domain, we do not need any further adaptation, the new domain can be used directly in the test phase. In contrast, DA methods need adaptation in the training for the new domain. 


DG is a less explored area of research which learns representations from all the available source domains. DG approaches can be broadly grouped into two categories: (1) shallow DG methods \cite{Muandet:2013:DGV:3042817.3042820,DBLP:conf/iccv/GhifaryKZB15,8053784,7437460} that use handcrafted features or deep features which are extracted using a pre-trained deep neural network (DNN); and (2) deep DG methods \cite{Li_2018_ECCV,Li_2018_CVPR,DBLP:journals/corr/abs-1807-08479,NIPS2018_7378,dou2019domain,li2019feature,li2019episodic,carlucci2019domain} where the model learns representations from the original sources images in an end-to-end manner using DNN. The deep features extracted from pre-trained DNNs are more descriptive and discriminative in nature compared to the handcrafted features. Thus, using deep features in DG has demonstrated the effectiveness in performance over using handcrafted features. 

Muandet et al. \cite{Muandet:2013:DGV:3042817.3042820} introduced a domain-invariant component analysis based DG approach for finding invariant transformation by decreasing the disparity among the source domains. Using scatter component analysis, Ghifary et al. \cite{DBLP:journals/pami/GhifaryBKZ17} introduced a unified framework which can be used for both DA and DG. Khosla et al. \cite{Khosla:2012:UDD:2402940.2402953} proposed a DG approach where the dataset shifts are decreased though the development of a dataset-specific model. To solve the DG problem, Ghifary et al. \cite{DBLP:conf/iccv/GhifaryKZB15} suggested a multi-task autoencoder approach for learning domain-invariant features. A deep DG method based on adversarial networks where the discrepancy among source domains is minimized by using class prior normalized and class conditional domain classification loss is proposed in \cite{Li_2018_ECCV}. Li et al.  \cite{DBLP:journals/corr/abs-1807-08479} introduced a DG framework where both marginal and conditional representations are considered. In \cite{8237853}, a DG framework is proposed based on a low rank parameterized convolutional neural network. Li et al. \cite{Li_2018_CVPR} proposed a MMD based adversarial autoencoder network to tackle the DG problem. Balaji et al.  \cite{NIPS2018_7378} proposed a DG model where the discrepancy is minimized using a regularization function.

Apart from shallow framework, the current DG approaches assume the same association of the labels of the source domains which is hardly practical in real-world scenarios. If this rule is violated i.e., if the source domains have uncommon categories, the model suffers poor performance due to negative transfer learning. Recent research \cite{Cao_2018_ECCV,2018_ECCV_Saito,open_2017_ICCV} of DA has examined the significance of the effect of negative transfer learning on the performance. To overcomes the above limitations of current approaches to DG, we propose an end-to-end feature-norm network  with collaborative learning, 
the details of which are explained in the following sections. 




\section{Proposed Approach}
\subsection{Problem Setup}

\begin{figure*}
\begin{center}
\includegraphics[width=1.0\linewidth]{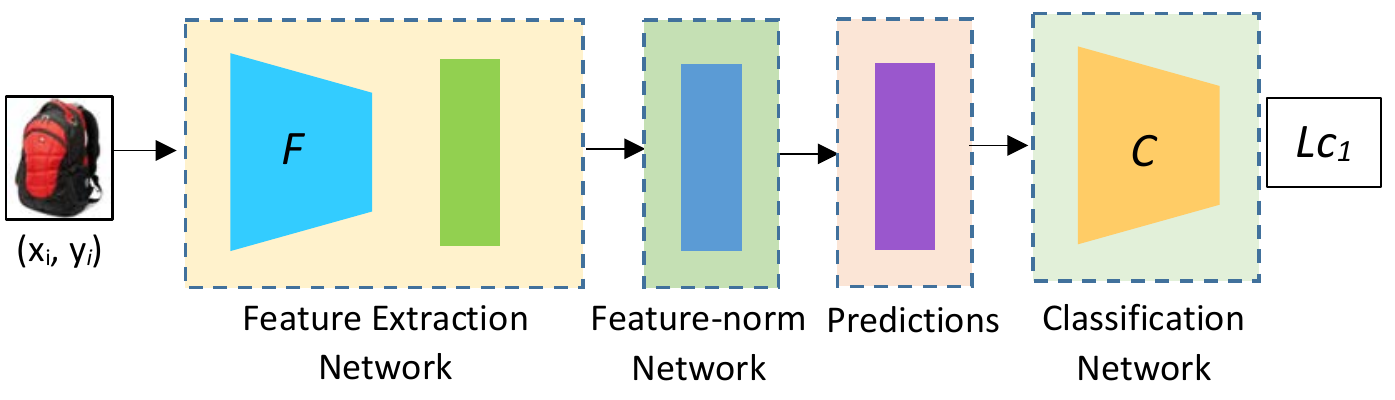}
\end{center}
   \caption{An overview of the proposed feature-norm network. The feature extraction module receives multi-source domains’ samples with their labels in the training phase and maps all the source data into the feature space. With regard to every individual example, we facilitate a feature-norm enlargement by the feature-norm network. The prediction network computes the predictions for each training sample and the classification network classifies the samples using supervised loss.
   }
\label{fig:architecture_fnn}
\end{figure*}

\begin{figure*}
\begin{center}
\includegraphics[width=0.9\linewidth]{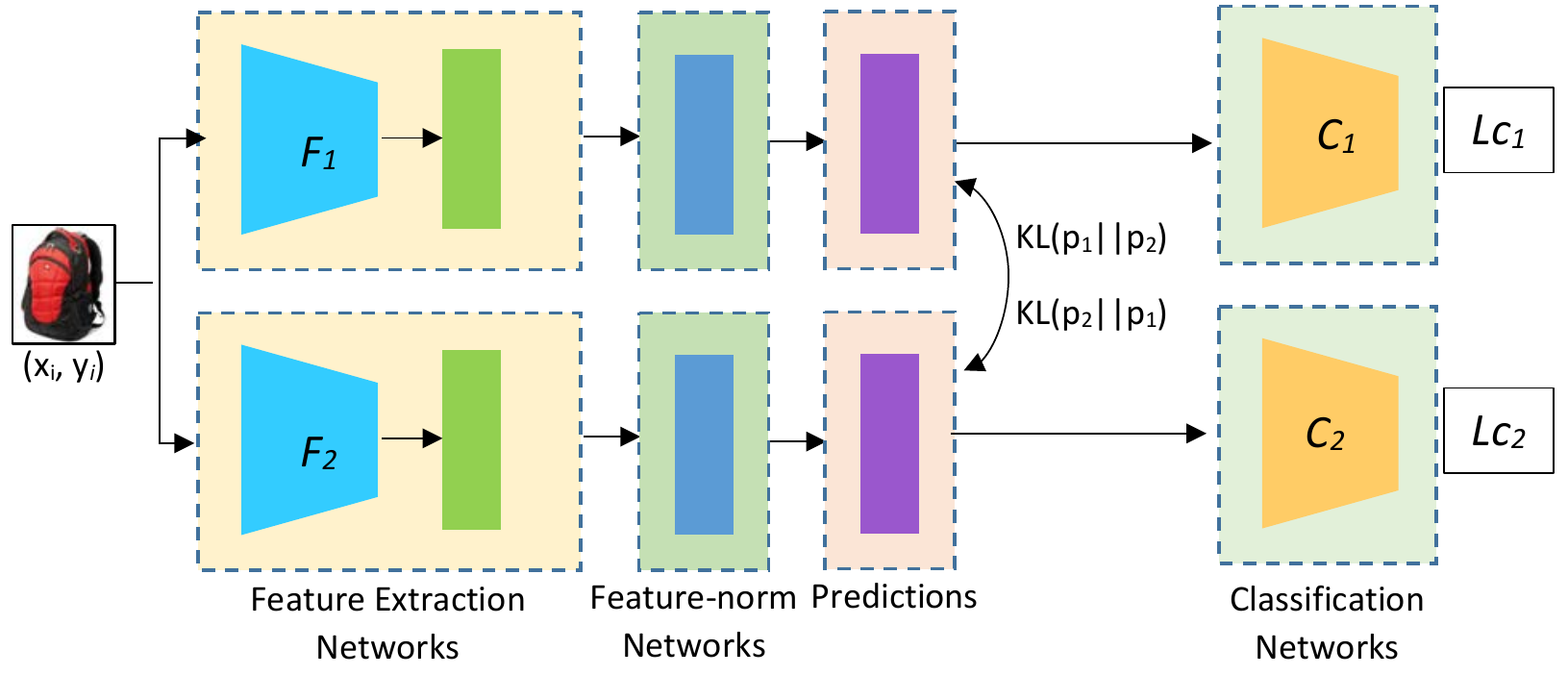}
\end{center}
   \caption{An overview of the proposed Collaborative Feature-norm Network (CFNN) for deep domain generalization. The feature extraction module receives multi-source domains' samples with their labels in the training phase and maps all the source data into the feature space. The feature-norm module receives the features from the source domains, then the feature-norm of the source domains are adapted to a shared scalar value $\Delta$r. A mimicry loss is used between the two networks prediction that aligns each network’s category posterior with the category probabilities of other networks. The prediction network computes the predictions for each training sample. The classification network classifies the samples using a supervised loss.
   }
\label{fig:architecture_cfnn}
\end{figure*}


We commence with a conventional representation of the DG problem. Suppose \( \mathcal{X} \) represents the feature space and \( \mathcal{Y} \) represents the label space. A domain \( \mathcal{D} \) is denoted by a joint distribution $P(X, Y)$ defined on  \( \mathcal{X \times Y} \). We assume that we have a set of $N$ number of source domains such as $\Omega$ = $S_D^1$; $S_D^2$; $S_D^3$;\dots;$S_D^N$ and target domain $T_D^N$ where $T_D^N$ $\notin$ $\Omega$. The goal of DG is to gain a classifying function $f: X \rightarrow Y$ able to classify $\{x_i\}$ to the corresponding $\{y_i\}$ given $S_D^1$; $S_D^2$; $S_D^3$;\dots;$S_D^N$ during training as the input, but $T_D^N$ is unavailable during the training phase. Previous DG methods assume all the source domains share the same label space $y_i$ = $y_j$, $\Delta$ =[\ 1, K]\ , where $\Delta$ is a set of classes. For more challenging setting, we raise the issue that the source domains may not have the same categories such as $y_i$ $\neq$ $y_j$. As the proposed method is independent of the association of the label space of the source domains, it can be effective in both the traditional setting as well as the category shift setting. We will start by discussing the traditional DG case first and will then introduce DG with category shift case later. 



\subsection{Feature-norm network (FNN)}
Figure \ref{fig:architecture_fnn}(a) shows the architecture of our proposed feature-norm network. It incorporates four parts: feature extraction network that depicts the invariant feature transformation, feature-norm network that reduces the domain discrepancy among the source domains, the prediction network that predicts the classes of each training image, and the image classification network that classifies the images using cross entropy loss. Feature extractor $F(.)$ is used to extract standard representations for the samples from all source domains. It takes the samples from source domains as input and maps the samples into a feature space. The aim of the feature-norm module is to obtain a domain agnostic model from the available source domains that can be applied to an unseen target domain. To tackle the negative transfer learning issue, we apply feature-norm adaptation to minimize the divergence among the source domains during training phase. Given a batch images $x_{i}$ from source domains, the common feature learning sub-network extract features $F(x_{i})$. Then, we apply feature-norm adaptation on the extracted features.

At each iteration, the feature norm adaptation is applied to the features with the supervised source classification loss. We promote the expansion of the feature-norm with respect to each example of the source domains at the step size of $\Delta r$. The domain shift among the source domains depends heavily on immoderately less informative features with smaller norms. We realize that more informative transferability is indicated by task-specific features of larger norms. We impose a larger-norm constraint on the task-specific features to facilitate the more insightful and transferable computation of the source domains. The prediction sub-network computes the predictions of the categories of each training instance. The classification sub-network $C$ is composed of a softmax classifier.

Now, we will briefly formalize the illustration above. Given $n$ number of samples from $N$ source domains $X$ = $\{x_{i}\}^{n}_{i=1}$ from $K$ categories, where the corresponding label is set as $Y$ = $\{y_{i}\}^{n}_{i=1}$ with $y_i \in \{1,2, \dots, K\}$. Each image from the source domains pass through the feature extraction network. In DG scenario, we assume that all the available source domains have labeled data. We use the supervised classification loss on the data from the source domains as follows,

\begin{align}\label{eq1}
\underset{F,C}{\min}  \sum_{j=1}^{N} E_{x_{\sim x_{i}}}  J (C(F(x_i^{})), y_i),
\end{align}

where $J(.,.)$ is the cross-entropy loss. The cross-entropy loss function can be formulated as,

\begin{align} \label{eq2}
L_{class} = - \sum_{i}^{K_{}} y_{i} \log (p_i),
\end{align}
where $K$ indicates the number of categories, $y_i$ indicates ground truth and $p_i$ indicates the predicted value. The predicted value of class $K$ for sample $x_i^{}$ given by a network $N_1$  is computed as,

\begin{align} \label{eq3}
p_i(x_i^{})= \frac{e^{z_{iN_1}}}{\sum^{K}_{k=1} e^{z_{kN_1}}},
\end{align}
where $e^{z_iN_1}$ is the logit calculated by the network $N_{1}$.

The simple approach for DG that aggregates all the source domains' data together and trains a single classifier ignoring the domain divergence among the source domains. The aggregation approach assumes the small divergence among the source domains. If the divergence is large, this approach will fail to make a good classifier for an unknown dataset due to the large divergence among the source domains. In real world situations, it's more obvious to obtain more diverse data where we can observe large divergence among the source domains. In DG the labeled data are collected from different diverse source domains and the characteristics of the data of these domains might be different from each other. Thus, we need to decrease the divergence among the source domains. The aim of DG approaches is to obtain domain-invariant representations from all the source domains. The straightforward technique of DG is to decrease the disparity between each pair of source domains in order to obtain domain-invariant representations in a common space for all the source domains. However, the common problem of defining such strategy is that it matches all the statistics of a source domain to other domains. As a result, if one domain has different categories to others, it will align the samples from the class to wrong classes that causes negative transfer learning. Hence, we introduce a DG framework which is independent of the association of the label space among the source domains to tackle the negative transfer learning. We construct an equilibrium and a large radius $R_{x_i}$ to bridge the gap among the source domains. For $N$ number of source domains such as, $S^{1}_{D}, S^{2}_{D} \cdots S^{N}_{D}$, the feature-norm objective is,

\begin{align} \label{eq7}
L_{domain} =  \gamma [ \frac{1}{n} ( \sum_{x_i \in S_D^N } {\|F(x_i)}\|_2 - R_{x_i})^2 ],
\end{align}

where $F{(x_i)}$ is the deep network feature for the sample $x_i$ from the source domain $N$, $R_{x_i}$ is the target norm value which can be learned, and $n$ is the total number of samples from domain $S^N_D$.

Equation \ref{eq7} restricts the mean feature-norms of the available source domains converging to $R_{x_i}$ to minimize the domain gap among source domains and gain a domain agnostic model that can be applied for the unseen target domain. In order to learn $R_{x_i}$ from the samples, we enlarge the feature-norm with respect to individual examples rather than a fixed value of $R_{x_i}$. During each iteration, Equation \ref{eq7} enlarges the feature-norm at $\Delta r$ step size according to the feature norms of the last iteration computed by the previous parameters of the model. The $R_{x_i}$ can be learned in the following way,
\begin{equation} \label{eq10}
R_{x_i} = \|F(x_{i})\|_2 + \Delta r,
\end{equation}
where $\Delta r$ represents the residual feature-norm which is used to control the expansion of the feature-norm.

We finally achieve the learning objective of FNN as follows,

\begin{align} \label{eq11}
L_{total} = L_{class} + L_{domain} 
= - \sum_{i}^{K_{}} y_{i} \log (p_i) +  \gamma [ \frac{1}{n} ( \sum_{x_i \in S_D^N } {\|F(x_i)}\|_2 - R_{x_i})^2 ].
\end{align}

The discrepancy of the available source domains predominantly stems from their much smaller feature norms. The proposed FNN for DG progressively generalize the feature norms of the source domains to a large range of values to obtain a domain agnostic model because the features with larger norms are more transferable.
\subsection{Collaborative Feature-norm Network (CFNN)}
Figure \ref{fig:architecture_fnn}(b) shows the proposed collaborative feature-norm network. In this approach, each image of every domain is passed through two networks and each domain is trained with three losses: supervised classification loss, domain discrepancy loss (feature-norm loss), and a mimicry loss that aligns each network's category posterior with the category probabilities of other networks. This collaborative learning allows each network to learn significantly better than when learning alone with the other two loss functions without mimicry loss. Though both networks predict the true class with the supervised classification loss for each training sample, the networks learn different representations as the networks start from a different initial condition. Thus the whole network pools the collective estimation of the next most similar categories. The most similar categories for each training samples according to their peers increases each networks posterior entropy that helps the network to converge to a more robust minimum with better generalization to target data in the test phase. Hence, we use another collaborative network to get the experience of dual learning in the form of the probability $p_i$ to increase the generalization capability of the network. To estimate the matching of the two network's probability we use mimicry loss (Kullback Leibler Divergence) as \cite{ying2018DML},
\begin{align} \label{eq1111}
D_{KL}(p_2||p1) = \sum_{i=1}^{n} \sum_{k=1}^{K} p_2^{k}(x_i) log \frac{p_2^{k}(x_i)}{p_1^k(x_i)}.
\end{align}

We finally achieve the learning objective of each network as follows to obtain the domain agnostic model,

\begin{equation} \label{eq6}
L_{total} = L_{class} + L_{domain} + D_{KL}(p_2||p_1).
\end{equation}

\subsection{Domain generalization with  category shift among source domains}
In DG, suppose we have $N$ number of source domains where the source domains have some common categories that exist in the target domain. The aim is to learn a domain agnostic model from the source domains that can be applied to the target data. As the domain discrepancy module is not dependent on the association of the labels of the source data, it can be directly used for DG with the category shift case. It is noteworthy that for category shifts in the source domains setting, the network architecture is still the same as the traditional setting of DG. We implement it with the same parameters and loss functions.

\section{Experiments}
In this section, we will discuss the experimental evaluation of our proposed approach for DG in both settings: traditional DG setting and DG with category shift setting. We evaluate our approach on three benchmark DG datasets to demonstrate the effectiveness of our method: PACS, Office-Home and office-Caltech datasets.

\subsection{Datasets} 
\textbf{PACS} \cite{8237853} is a recently proposed benchmark for DG task which is created by considering the common classes among Caltech256 , Sketchy, TU-Berlin and Google Images. This dataset consists of 4 domains - Art painting, Cartoon, Photo and Sketch. It has total $9991$ images with $7$ shared classes: dog, guitar, giraffe, elephant, person, horse, house. We evaluate our proposed method on four different transfer tasks P, C, S $\rightarrow$ A; P, A, S $\rightarrow$ C; A, C, S $\rightarrow$ P; and A, C, P $\rightarrow$ P. The transfer task P, C, S $\rightarrow$ A indicates three source domains Photo (P), Cartoon (C) and Sketch (S) and one target domain Art-Painting (A). We follow the standard protocol for domain generalization where during the training phase we access the labeled source data but not access the target data. The target data is used only in test phase only.  

\textbf{ Office-Caltech} \cite{DBLP:conf/cvpr/GongSSG12} dataset is formed by taking the 10 common classes of two datasets: Office-31 and Caltech-256, which is a standard benchmark for visual object classification task. Each class has 8 to 151 images and in total it consists of 2533 images. It has 4 domains: Amazon (A), Webcam (W), DSLR (D) and Caltech (C). We evaluate our proposed method on ten transfer tasks C, D, W $\rightarrow$ A; A, D, W $\rightarrow$ C; C, A, W $\rightarrow$ D; A, C, D $\rightarrow$ W; A, C$\rightarrow$ D,W; DW$\rightarrow$ A, C; A, W $\rightarrow$ C, D; A, D$\rightarrow$ C, W; C, W$\rightarrow$ A, D and C, D$\rightarrow$ A, W The transfer task C, P, R $\rightarrow$ A indicates there are three source domains Clipart (C), Product (P) and Real-world (R); and one target domain Art (A).

\textbf{Office-Home} \cite{venkateswara2017Deep} is a recently proposed dataset for DA where the images are collected from $4$ different domains: Real-World (Rw) images, Artistic images (Ar), product images (Pr) and clipart images (Cl). Each domain consists of 65 object classes. It contains a total of around 15500 images.  We evaluate our proposed method on four transfer tasks C, P, R $\rightarrow$ A; A, P, R $\rightarrow$ C; C, A, R $\rightarrow$ P; and C, P, A $\rightarrow$ R. The transfer task C, P, R $\rightarrow$ A indicates there are three source domains Clipart (C), Product (P) and Real-world (R); and one target domain Art (A).

\begin{table*}[h!]
\centering
\small\addtolength{\tabcolsep}{11.5pt}
 \caption{Recognition accuracies for DG on the PACS dataset using pretrained AlexNet.}
 \resizebox{14.5cm}{!}{
  \begin{tabular}{|c | c  c c c c|} 
 \hline
 Source $\rightarrow$ Target  & A & C &  P &  S  &Ave. \\ [0.5ex] 
  \hline
  
  SCA \cite{DBLP:journals/pami/GhifaryBKZ17} &50.05 &58.79 &59.10 &50.62 &54.64 \\
  
  MTAE \cite{DBLP:conf/iccv/GhifaryKZB15} &45.95 &51.11 &58.44 &49.25 &51.19 \\
  
  Source only &64.91 &64.28 &86.67 &53.08 &67.24 \\
  
  CIDDG \cite{Li_2018_ECCV} &62.70 &69.73 &78.65 &\textbf{64.45} &68.88 \\
  
  DBADG \cite{8237853}  &62.86 &66.97 &89.50 &57.51 &69.21 \\
  
  DSN \cite{NIPS2016_6254} &61.13 &66.54 &83.25 &58.58 &67.4 \\
  
  MLDG \cite{li2018learning} &66.23 &66.88 &88.00 &58.96 &70.02 \\
  
   \hline
  \textbf{FNN (Ours)} &67.31 &70.49 &90.67 &60.53 &72.53 \\
  
 \textbf{CFNN (Ours)} &\textbf{69.53} &\textbf{72.32} &\textbf{91.56} &63.19&\textbf{74.15} \\
  
 
 \hline
 \end{tabular}}
\label{table_pacs_AlexNet}
\end{table*}

\begin{table*}[h!]
\centering
\small\addtolength{\tabcolsep}{12.5pt}
 \caption{Recognition accuracies for DG on the PACS dataset using pretrained ResNet-50.}

  \begin{tabular}{|c |  c  c c c c|} 
 \hline
 Source $\rightarrow$ Target  & A &C &P &S  &Ave. \\ [0.5ex] 
  \hline
  
  Source only &84.76 &76.27 &93.29 &70.14 &81.12  \\
  
  CIDDG \cite{Li_2018_ECCV} &87.40 &77.12 &94.52 &73.28 &83.08 \\
  
  
  
  MLDG \cite{li2018learning} &81.21 &77.46 &94.82 &72.64 &81.78 \\
  
  \hline
  \textbf{FNN(Ours)} &86.71 &78.49 &95.38 &74.94 &83.88 \\
  
    \textbf{CFNN(Ours)} &\textbf{88.26} &\textbf{81.10} &\textbf{96.40} &\textbf{76.55} &\textbf{85.59}  \\
  
 
 \hline
 \end{tabular}
\label{table_pacs_ResNet}
\end{table*}

\begin{table*}[h!]
\centering
\small\addtolength{\tabcolsep}{11.9pt}
 \caption{Recognition accuracies for DG on the Office-Caltech dataset  using pretrained AlexNet.}
 \begin{tabular}{|c |  c c c c c |} 

 \hline
 Source $\rightarrow$ Target  & A &C &D & W  &Ave. \\ [0.5ex] 
  \hline
  
  KPCA \cite{Scholkopf:1998:NCA:295919.295960} &90.92 &74.23 &94.34 &88.84 &87.08 \\
  
  DICA \cite{Muandet:2013:DGV:3042817.3042820} &80.34 &64.55 &93.21 &69.68 &76.95 \\
  
  Undo-Bias \cite{Khosla:2012:UDD:2402940.2402953} &89.56 &82.27 &95.28 &90.18 &89.32 \\
  
  SCA \cite{DBLP:journals/pami/GhifaryBKZ17} &89.97 &77.90 &93.21 &81.26 &85.59 \\
  
  
  Source only  &89.35 &83.79 &97.72 &93.88 &91.19 \\
  
  
  CIDG \cite{DBLP:journals/corr/abs-1807-08479} &93.24 &85.07 &97.36 &90.53 &94.05 \\
 \hline
 
  \textbf{FNN(Ours)} &92.35 &86.79 &100.00 &95.27 &93.60 \\

  \textbf{CFNN(Ours)} &\textbf{94.19} &\textbf{87.29} &\textbf{100.00} &\textbf{96.76} &\textbf{94.56} \\
 
 \hline
 \end{tabular}
\label{off_call_alex_3source}
\end{table*}

\begin{table*}[h!]
\centering
\small\addtolength{\tabcolsep}{5pt}
 \caption{Recognition accuracies for DG on the Office-Caltech dataset using pretrained AlexNet.}
  \begin{tabular}{|c |  c c c c c c c|} 
 \hline
 Source $\rightarrow$ Target  &D,W & A,C  &C,D & C,W & A, D &A,W   &Ave. \\ [0.5ex] 
  \hline
  
  KPCA  \cite{Scholkopf:1998:NCA:295919.295960} &75.81 &65.75 &76.26 &75.81 &91.45 &90.36 &79.24 \\
  
  DICA \cite{Muandet:2013:DGV:3042817.3042820} &60.41 &43.02 &69.29 &68.49 &83.01 &79.69 &67.32 \\
  
  Undo-Bias \cite{Khosla:2012:UDD:2402940.2402953} &80.24 &74.14 &81.77 &81.23 &91.73 &90.67 &83.30 \\
  
  SCA \cite{DBLP:journals/pami/GhifaryBKZ17} &76.89 &69.53 &78.99 &75.84 &90.46 &88.61 &80.05 \\
  
  
  Source only  &80.75 &74.67 &85.85 &86.05 &91.01 &91.85 &85.03 \\
  
  
  CIDG \cite{DBLP:journals/corr/abs-1807-08479} &83.65 &65.91 &83.89 &84.66 &93.41 &91.70 &83.87 \\
 \hline
 
 \textbf{FNN (Ours)} &83.93 &76.58 &87.12 &88.37 &94.30 &93.74 &87.34 \\

  \textbf{CFNN(Ours)}&\textbf{85.51} &\textbf{78.33} &\textbf{88.87} &\textbf{89.68} &\textbf{95.92} &\textbf{95.41} &\textbf{88.95} \\
  
 
 \hline
 \end{tabular}
\label{off_call_alex_2source}
\end{table*}

\begin{table*}[h!]
\centering
\small\addtolength{\tabcolsep}{12pt}
 \caption{Recognition accuracies for DG on the Office-Home dataset using pre-trained ResNet-50.}
  \begin{tabular}{|c |  c  c c c c|} 
 \hline
 Source $\rightarrow$ Target  &Ar & Rw &Cl & Pr  &Ave. \\ [0.5ex] 
  \hline
  
  Source only &66.29 &81.25 &51.61 &79.25 &69.60  \\
  
  CIDDG \cite{Li_2018_ECCV} &67.56 &82.87 &53.18 &80.69 &71.07 \\
  CIDG \cite{DBLP:journals/corr/abs-1807-08479} &67.85 &82.59 &52.91 &80.37 &70.93 \\
  DBADG \cite{8237853} &66.43 &81.51 &51.38 &79.82 &69.79 \\
  
  \hline
 
  \textbf{FNN(Ours)} &68.34 &82.94 &53.41 &80.12 &71.20 \\ 
  
  \textbf{CFNN(Ours)} &\textbf{69.76} &\textbf{83.56} &\textbf{54.18} &\textbf{81.64} &\textbf{72.29}  \\
  
 
 \hline
 \end{tabular}
\label{table_off_home_ResNet}
\end{table*}

\begin{table*}[h!]
\centering

\small\addtolength{\tabcolsep}{4pt}
 \caption{For $Rw,Cl,Pr \rightarrow Ar$ and $Ar,Cl,Pr \rightarrow Rw$ transfer tasks on the Office-Home dataset using pre-trained ResNet-50 in the category shift setting.}

\begin{tabular}{|p{2cm}|p{1.2cm}|p{1.4cm}|p{1.2cm}||p{1.2cm}|p{1.4cm}|p{1.2cm}|}
 \hline
 \multirow{3}{*}{Models} & \multicolumn{3}{c|}{$Rw,Cl,Pr \rightarrow Ar$} & %
    \multicolumn{3}{c|}{$Ar,Cl,Pr \rightarrow Rw$} \\
    \cline{2-7}
    & Accuracy & Degraded Accuracy &Transfer Gain  & Accuracy & Degraded Accuracy &Transfer Gain \\
  \hline
  
  Source only &62.70 &-3.59 &0 &76.98 &-4.27 & 0   \\
  CIDDG \cite{Li_2018_ECCV} &64.32 &-3.24 &1.62 &77.52 &-5.35 &0.54   \\
  
  CIDG \cite{DBLP:journals/corr/abs-1807-08479} &63.58 &-4.27 &0.88 &77.16 &-5.54 &0.18    \\
  
  DBADG \cite{8237853} &62.19 &-4.24 &-0.51 &76.85 &-4.66 &-0.13   \\
  
  \hline
  \textbf{FNN(Ours)} &66.13 &-2.21 &3.43 &81.35 &-1.59 &4.37   \\
  
 \textbf{CFNN(Ours)} &\textbf{67.57} &\textbf{-2.19} &\textbf{4.87} &\textbf{82.10} &\textbf{-1.46} &\textbf{5.12}     \\
  
 
 \hline
 \end{tabular}
\label{table_off_home_ResNet_partial}
\end{table*}

 \begin{figure*}
\begin{center}
\includegraphics[width=1.0\linewidth]{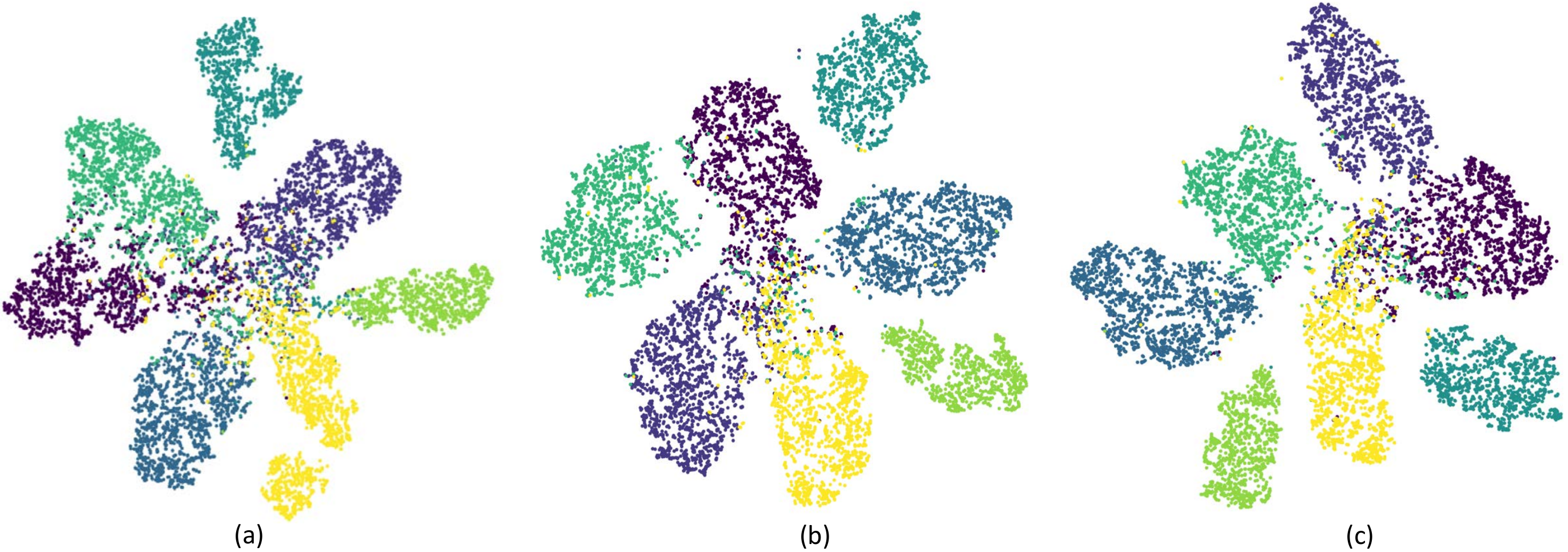}
\end{center}
   \caption{t-SNE visualization of the activations of (a) source only (b) FNN (c) CFNN on P, A, S $\rightarrow$ C transfer task of PACS dataset.}
   
\label{fig:tsne_f}
\end{figure*}

\subsection{Comparison with state-of-the-art}
We compare our proposed approach with several state-of-the-art DG methods. \textbf{Undo-Bias} \cite{Khosla:2012:UDD:2402940.2402953} is a multi-task support vector machine based model to address domain shift utilizing a domain specific and a global weight. \textbf{SCA} \cite{DBLP:journals/pami/GhifaryBKZ17} is a unified model for DA and DG where the discriminability of categories are maximized and the divergence of the domains are minimized. \textbf{DICA} \cite{Muandet:2013:DGV:3042817.3042820} stands for domain invariant component analysis which learns a domain-invariant feature representation. \textbf{KPCA} (kernel principal component analysis) \cite{Scholkopf:1998:NCA:295919.295960} discovers the superior components of the extracted features.  \textbf{MTAE} \cite{DBLP:conf/iccv/GhifaryKZB15} is a DG approach where the original images from the source domains are transformed into analogs in multiple related domains. \textbf{Source only} is the simple source domains aggregation approach for DG without any adaptive loss function. \textbf{DSN} \cite{NIPS2016_6254} stands for domain separation networks that decrease the divergence among source domains using shared and private spaces. \textbf{CIDDG} \cite{Li_2018_ECCV} is a deep DG method based on adversarial networks where the discrepancy among source domains is minimized by using class prior normalized and class conditional domain classification loss. \textbf{MLDG} \cite{li2018learning} is a meta-learning based DG method. \textbf{CIDG} \cite{DBLP:journals/corr/abs-1807-08479} is a DG framework where both marginal and conditional representations are considered to mitigate the DA problem. \textbf{DBADG} \cite{8237853} is a DG framework based on low rank parameterized convolutional neural network. 


\subsubsection{Settings}
We follow the same settings as CIDDG, CIDG and DBADG for the traditional DG task where we assume that all the source domains have the same categories. On the other hand, in DG with the category shift setting, we assume that all the source domains have a different label space. As our DG approach is not dependent on the association of the labels of the data, it can be directly applied to this category shift setting.


\subsection{Implementation details}

We implemented our method on the PyTorch \footnote{https://pytorch.org/} deep-learning framework. Our backbone networks are ResNet-50 and AlexNet, and we finetuned from the ImageNet \cite{NIPS2012_4824} pre-trained model. We used mini-batch stochastic gradient descent (SGD) with the momentum of 0.9 and learning rate $1 \times 10^{-3}$ for both backbone networks. We adopted a unified set of
hyper-parameters for all experiments, where $\gamma$ = 0.05, $\Delta r$ = 1.0. We repeated each
experiment five times and reported the average accuracy with the standard deviation. For FNN, we used either ResNet-50 or AlexNet network. For CFNN, we used the same architecture for two streams network. The model is scalable to a large number of source domains as we used pre-trained network. Our code will be made publicly available at time of camera-ready submission. 

\subsection{Results}

We test the DG performance of our proposed method in traditional settings, and compare the results with several state-of-the-art approaches, reported in Table \ref{table_pacs_AlexNet} to \ref{table_off_home_ResNet}. We also test our approach in the DG with category shift setting. The results are reported in Table \ref{table_off_home_ResNet_partial}.
The experimental results on traditional DG settings on PACS dataset are reported in Table \ref{table_pacs_AlexNet} using AlexNet as pre-trained model. From the results, we can see that our proposed CFNN performs the best 3 of the 4 DG transfer tasks. Our proposed FNN and CFNN maintains the best performance overall, with a $3-4$ \% and $5-6$ \% improvement on the source domain aggregation (Source only) model respectively.

The results are reported in Table \ref{off_call_alex_3source} for the Office-Caltech dataset. From the results, we can see that the previous state-of-the-art method was achieved by \cite{DBLP:journals/corr/abs-1807-08479} where the discrepancy among source domains is minimized based on domain-invariant class-conditional distributions. In contrast, CFNN achieves a domain invariant model using the feature-norm technique and achieves 94.56\% average recognition accuracy which is a significant improvement in performance. It is also noted that CFNN outperforms FNN significantly.

For the Office-Home dataset, the results are reported in Table \ref{table_off_home_ResNet} using ResNet as a pre-trained  model in traditional DG settings. Our method outperforms the previous state-of-the-art method \cite{Li_2018_ECCV} by 1.2\% average accuracy over the four transfer tasks. The difference with our work is that \cite{Li_2018_ECCV} used adversarial learning to reduce the divergence among source domains whereas we used feature-norm technique to minimize the domain discrepancy. In this dataset, our CFNN model is performing better than our FNN approach.

From the above evaluation, we can make a prominent conclusion that the traditional DG methods (shallow architecture) perform even worse than the Source only method, which fine-tunes the model using all source domains. Deep DG methods perform better than the Source only method. Moreover, our proposed approach for end-to-end deep DG performs better than the Source only and other DG approaches and sets a new state-of-the-art performance.

In the category shift setting of DG, we evaluated our model on the Office-Home dataset in two transfer tasks that are reported in Table \ref{table_off_home_ResNet_partial}. In this setting, we randomly selected 10 categories among 65 which are uncommon across the source domains. The accuracy deterioration when it is compared to the traditional DG setting and the transfer gain when it is compared to Source only approach are also appended. According to the results, CIDG suffers negative transfer gain mostly. In comparison to the traditional DG  approaches, FNN and CFNN mitigate the efficiency degradation problem and achieve positive transfer gains in all cases. This shows that FNN and CFNN will prevent the negative transfer between the source domains induced by the category change.

\section{Visualization}

 A t-SNE visualization of the learned embeddings is performed on the $P,A,S \rightarrow C$ transfer task of the PACS dataset to further evaluate the efficacy of our proposed approach as shown in Figure \ref{fig:tsne_f}. From the Figure  \ref{fig:tsne_f}, FNN and CFNN optimizes the embedding space such that the data points with the same categories are closer to each other than source only. However, CFNN decreases the discrepancy among the source and target domain more than FNN.

\section{Sensitivity of $\Delta r$}

We conduct case studies to analyse the differences of the norm values in our FNN and CFNN for the Rw, Cl, P r $\rightarrow$ Ar and Ar, Cl, Pr $\rightarrow$ Rw transfer tasks on Office–Home dataset. The results are shown in the Table \ref{table_sensitivity_r} by varying $\Delta r$. From the results, it can be seen that the performance remains almost stable as the parameter $\Delta r$ varies, suggesting that FNN and CFNN considerately stable on these two transfer tasks. 

\begin{table*}[h!]
\fontsize{6.6}{8}\selectfont 
\centering
 \caption{Sensitivity of  norm value.}
  \begin{tabular}{|c |  c | c | c |c |} 
 \hline
 $\Delta r$ & Rw, Cl, Pr $\rightarrow$ Ar (FNN) & Rw, Cl, Pr $\rightarrow$ Ar (CFNN) & Ar, Cl, Pr $\rightarrow$ Rw (FNN) & Ar, Cl, Pr $\rightarrow$ Rw (CFNN) \\ [0.5ex] 
  \hline
  
 0.5 &67.29  &68.38  &82.06  &82.74 \\
 1.0 &68.34 &69.76  &82.94  &83.56     \\
 1.5 &67.51  &68.40  &82.21 & 82.91    \\
 
 \hline
 \end{tabular}
\label{table_sensitivity_r}
\end{table*}

\section{Conclusion}

In this paper, we have presented an end-to-end deep domain generalization framework. We have achieved domain generalization by enlarging the feature norm of the samples. Hence, our proposed approach is independent of the association of the classes among the source domains so that the proposed method can be applied to both traditional domain generalization and category shift domain generalization settings. We have also improved our feature norm domain generalization model by introducing collaborative learning which enhances the generalization capability to the test samples using the predictions of the next most likely categories. We have demonstrated the effectiveness of the proposed approach on several benchmarks and have exceeded the state-of-the-art performance in most of the transfer tasks in the both domain generalization settings.  

\section*{Acknowledgements}
The research presented in this paper was supported by Australian Research Council (ARC) Discovery Project Grant
DP170100632.

\bibliography{mybibfile}

\end{document}